\RequirePackage{fix-cm}
\documentclass{svjour3}                     
\smartqed  
\usepackage[utf8]{inputenc}
\usepackage[english]{babel}
\usepackage{natbib}
\usepackage{graphicx}
\usepackage{amsmath, amssymb, dsfont, bm}
\usepackage{hyperref}
\usepackage{mathptmx}
\usepackage[english, ruled]{algorithm2e}
\begin{document}

\title{Improving Reliability of Latent Dirichlet Allocation by Assessing Its Stability Using Clustering Techniques on Replicated Runs
}
\titlerunning{Improving Reliability of LDA by Assessing Its Stability}        

\author{Jonas Rieger \and Lars Koppers \and Carsten Jentsch \and Jörg Rahnenführer}


\institute{Jonas Rieger ORCiD: 0000-0002-1642-9616 \at
              Department of Statistics, TU Dortmund University, 44221 Dortmund, Germany \\
              Tel.: +49 231-755 5419\\
              \email{rieger@statistik.tu-dortmund.de}
           \and
           Lars Koppers ORCiD: 0000-0002-0007-4478 \at Department of Statistics, TU Dortmund University, 44221 Dortmund, Germany \\
           \emph{Present address:} Institute of Technology Futures, KIT, 76133 Karlsruhe, Germany
            \and Carsten Jentsch ORCiD: 0000-0001-7824-1697 \and Jörg Rahnenführer ORCiD: 0000-0002-8947-440X
         \at Department of Statistics, TU Dortmund University, 44221 Dortmund, Germany
}


\maketitle

\begin{abstract}
For organizing large text corpora topic modeling provides useful tools. A widely used method is Latent Dirichlet Allocation (LDA), a generative probabilistic model which models single texts in a collection of texts as mixtures of latent topics. In this approach each topic is characterized by its word distribution, which is obtained by assigning words to topics. The assignments of words to topics rely on initial values such that generally the outcome of LDA is to some extent random and therefore not fully reproducible. In addition, the reassignment via Gibbs Sampling is based on conditional distributions, leading to different results in replicated runs on the same text data. This fact is often neglected in everyday practice.
We aim to improve the reliability of LDA results. Therefore, we study the stability of LDA by comparing assignments from replicated runs. We propose to quantify the similarity of two generated topics by a modified Jaccard coefficient. Using such similarities, topics can be clustered. 
A new pruning algorithm for hierarchical clustering results based on the idea that two LDA runs create pairs of similar topics is proposed. This approach leads to the new measure S-CLOP ({\bf S}imilarity of multiple sets by {\bf C}lustering with {\bf LO}cal {\bf P}runing) for quantifying the stability of LDA models.
We discuss some characteristics of this measure and illustrate it with an application to real data consisting of newspaper articles from \textit{USA Today}.
Our results show that the measure S-CLOP is useful for assessing the stability of LDA models or any other topic modeling procedure that characterize its topics by word distributions. Based on the newly proposed measure for LDA stability, we propose a method to increase the reliability and hence to improve the reproducibility of empirical findings based on topic modeling. This increase in reliability is obtained by running the LDA several times and taking as prototype the most representative run, that is the LDA run with highest average similarity to all other runs.
\end{abstract}

\section{Introduction}
\label{introduction}
Understanding unstructured data, e.g. texts, is a big challenge, especially due to the complicated and not always standardized structure and the increasingly large volumes of text data. Text data are the most frequent data type in the world today \citep{rajaraman2016}, and text mining tools have become very popular to analyze such data.

Text data are usually organized in large corpora, where each corpus consists of a collection of $M$ texts, often also denoted as documents or articles. Each text can be considered as a sequence of tokens of words of length $N^{(m)}, m = 1, ..., M$, where $N^{(m)}$ is the total number of tokens used in text $m$. In common notation token means an individual word at a specific place in the text and the set of words is used synonymously with vocabulary. Then, the number of different words used in a text corpus is denoted by $V$. We refer to these terms in the following.

\subsection{The Latent Dirichlet Allocation and its weaknesses}
\label{lda}
Topic modeling \citep{blei2012} is one popular method of text mining, in particular Latent Dirichlet Allocation (LDA) \citep{blei2003} that assumes distributions of latent topics for each text. Topics are referred to as $T_n^{(m)}$ for the topic assignment of token $w_n^{(m)}$ at position $n = 1, ..., N^{(m)}$ in text $m$, with $K$ topics in total. The set of possible topics is given by $\bm Z = \{Z_1, ..., Z_K\}$, so that $T_n^{(m)} \in \bm Z$. Then, the count of assignments of a word $v = 1, ..., V$ to topic $Z_k$ is denoted by $n_k^{(v)} \in \mathbb{N}_0$, and the counts of assigned words can be summarized in the vector
\begin{align*}
\bm z_k = \left(n_{k}^{(1)}, ..., n_{k}^{(V)} \right)^T \in \mathbb{N}_{0}^V,
\end{align*}
so that the matrix of word counts per topic is given by $\bm z = \left(\bm z_1, ..., \bm z_K \right) \in \mathbb{N}_0^{V \times K}$. In a seminal paper \citep{blei2003} the method of LDA has been proposed to obtain such word counts per topic. The probability model of LDA \citep{griffiths2004} is given by 
\begin{align*}
&w_n^{(m)} \mid T_n^{(m)}, \bm\phi_k & \sim~~& \text{Discrete}(\bm\phi_k),&\\
&\bm\phi_k & \sim~~& \text{Dirichlet}(\beta),&\\
&T_n^{(m)} \mid \bm\theta_m & \sim~~& \text{Discrete}(\bm\theta_m),&\\
&\bm\theta_m & \sim ~~& \text{Dirichlet}(\alpha).&
\end{align*}

For a given parameter set $\{K, \alpha, \beta\}$, LDA assigns one of the $K$ topics to each token. Here $K$ denotes the number of topics and $\alpha, \beta$ are parameters of a Dirichlet distribution defining the type of mixture of topics in one text and the type of mixture of words in one topic. Higher values for $\alpha$ lead to a more heterogeneous mixture of topics whereas lower values are more likely to produce less but more dominant topics per text. Analogously, $\beta$ controls the mixture of words in topics. Topic distributions per text $\bm\theta_m$ and word distributions per topic $\bm\phi_k$ are (often) estimated using a Collapsed Gibbs Sampler \citep{griffiths2004}. In this procedure the initial assignment of a token is random and its reassignment is based on the conditional distributions, which leads to different results in multiple LDA runs for fixed parameters.

This instability of LDA leads to a lack of reliability of the modeling results. Reliability needs to be quantified to judge text mining findings using LDA models. In this paper, we propose a novel measure for this purpose. In a concrete example we show that single models are prone to misinterpretation because of the large difference in results from LDA runs. To overcome this issue we propose a method for increasing the reliability of LDA runs by determining a prototype model that is most representative, in the sense that it is most similar to other runs taken from the same modeling procedure. In the following we refer to this most representative model as prototype. A few approaches exist to make LDA results more reliable, but all of them have weaknesses. Often the modeling procedure itself is influenced in a way such that LDA loses its flexibility. Other methods do not search in the whole space of possible models, which leads to non-optimal results. To address these weaknesses, our approach does not affect the modeling procedure and is exclusively based on replicated runs.

\subsection{Methods and modifications of LDA to overcome the instability}
\label{modifications}
\cite{agrawal2018} point out that instability is mentioned by most of the recent works on LDA, though all of them use the default parameters and only a few of them tune the model. They propose a new algorithm LDADE (LDA \textbf{D}ifferential \textbf{E}volution) which automatically tunes the parameters of LDA in order to optimize topic similarity in replications using a differential evolution algorithm. This results in a set of input parameters $K, \alpha$ and $\beta$ which perform best on the given data with respect to model stability. This procedure does not really increase the reliability for a given parameter set, but tries to find the best parameter set regarding a stability score. The implicit parameter optimization of the mentioned procedure can make one believe that the resulting model is optimal. However, independent of the underlying data set, specific parameters could result in systematically better stability values. In addition, for LDADE the goodness of the fit regarding interpretability is not validated. 

In contrast, \cite{maier2018} aim for increasing both, reliability and interpretability of the final model simultaneously. Therefore, they maximize topic similarity as well as topic coherence, but discover that standard metrics do not perform well in increasing interpretability in general. Instead, manual approaches as e.g.\ the intruder validation technique proposed by \cite{chang2009} are essential. \cite{maier2018} propose to increase reliability of LDA by restricting the set of possible outcomes with initialization techniques.

There is also a modification of the LDA implementation that reduces instability. GLDA (\textbf{G}ranulated LDA) was proposed by \cite{koltcov2016} and is based on a modified Gibbs Sampler. The idea of the algorithm is that tokens $w_n^{(m)}$ that are closer to each other are more likely to be assigned to the same topic. The authors show that their algorithm performs comparably well with standard LDA regarding interpretability. Moreover, it leads to more stable results. Their study is based only on three LDA runs and the implementation is not publicly available. Thus, a validation of this method on other datasets or with larger numbers of replications is pending.

\subsection{Similarity measures for topics}
\label{measures}
For quantifying similarities between (LDA) models, it is necessary to first determine similarities between topics. A similarity between two topics can be calculated based on the two corresponding vectors of counts. We build on the well established Jaccard coefficient \citep{jaccard1912} and introduce a new modification. Its general form is given by
\begin{align*}
\text{Jaccard}(A, B) =  \frac{\vert A \cap B \vert}{\vert A \cup B \vert},
\end{align*}
where $A, B$ are sets of words. In our modification of the Jaccard coefficient, introduced in Section \ref{methods}, we will restrict the sets $A$ and $B$ to words that are assigned more often than a given threshold to the two corresponding topics.

While \cite{agrawal2018} determine topic similarity with a Jaccard coefficient of the top $9$ words per topic across multiple runs and measure stability with the median of the topic similarities, \cite{maier2018} use the cosine similarity
\begin{align*}
\text{cosine}(a, b) = \frac{a \cdot b}{\Vert a \Vert_2 \cdot \Vert b \Vert_2},
\end{align*}
where $a, b$ are word counts. For repetitions of the same modeling procedure they match topics with the highest cosine similarity, which additionally has to be greater than the threshold~$0.7$. Then, for two models the similarity is calculated as the share of topic matches, and for more than two models by the mean of all pairwise shares.

\cite{greene2014} and \cite{su2016} determine topic similarities with an average Jaccard coefficient
\begin{align*}
\text{AverageJaccard}(A, B) = \frac{1}{N}\sum_{n = 1}^N \frac{\vert A_n \cap B_n \vert}{\vert A_n \cup B_n \vert},
\end{align*}
where $A_n$ and $B_n$ define the sets of the first $n$ words of the ordered lists from the word sets A and B. They choose $n = 5$ and find the best matching topics of different LDA runs based on this measure with the hungarian method \citep{kuhn1955}. The authors try to encounter the problem that more than two runs of topics have to be matched, by learning a reference model. They calculate the similarity of one LDA to the reference LDA as the mean average Jaccard coefficient over all matched topics $Z_{k^*}$ to the topics  $Z_k^{(\text{ref})}$ of the reference model. Here $Z_{k^*}$ denotes the reordered topics $Z_{k}$ of the LDA run, so that matching topics have the same index. Analogously, they calculate stability over a number of $R$ replications as the mean over the pairwise similarities against the (predetermined) reference model $\bm Z^{(\text{ref})}$ by
\begin{align*}
\frac{1}{R} \sum_{r=1}^R \left( \frac{1}{K} \sum_{k=1}^K \text{AverageJaccard}\left( Z_k^{(\text{ref})}, Z_{k^*}^{(r)}\right) \right).
\end{align*}
One drawback of this approach is the specification of the reference model, which should be a good representative of the other LDAs. It is non-trivial to determine this representative model. Therefore, our approach follows an opposite strategy. We first calculate similarities between models and then determine the prototype model based on these values.

Another option for measuring topic similarity introduced by \cite{mantyla2018} is the Rank Biased Overlap (RBO) \citep{webber2010} for comparing ranked lists. A parameter controls how much influence the word order has. While the measure seems to be useful because it implements a more flexible form of a Jaccard coefficient, the authors do not investigate stability of LDA models based on RBO. Moreover, the calculation of the measure is very time consuming, also due to the required parameter optimization.

In this work we propose to assess the stability of LDA with clustering techniques applied to replicated LDA runs. High stability corresponds to high reliability of findings based on stable models in the sense of improving reproducibility. We introduce a new automated method of clustering topics, more precisely a pruning algorithm for results of hierarchical clustering, based on the optimality criterion that each cluster should contain one topic of every replication of the modeling procedure. This results in our novel similarity measure S-CLOP ({\bf S}imilarity of multiple sets by {\bf C}lustering with {\bf LO}cal {\bf P}runing) for multiple sets of objects. In the given application example the multiple sets are LDA runs and the objects are topics. We demonstrate the potential of this measure to quantify reliability by applying it to an example corpus from the daily American newspaper \textit{USA Today}. This corpus contains $M = 7\,657$ articles from June until November 2016. We propose a repetition strategy to increase the reliability of findings from topic models using our novel measure S-CLOP to calculate similarities of topic models.

\section{Methods}
\label{methods}
For assessing LDA stability, an adequate similarity measure for topics is required. We define a Jaccard coefficient which is modified in the sense that not all words are considered as relevant for each topic. Further, we introduce our novel similarity measure S-CLOP for topic models, which is the basis for quantifying stability of LDA. In Section \ref{results}, the new measure S-CLOP is applied to LDA replications on an example corpus and is used to increase reliability of the findings from an LDA model.

\subsection{Modified Jaccard coefficient: a similarity measure for topics}
\label{topicsim}
Suppose we have a text corpus and we estimate a topic model with LDA, with parameters $\alpha$, $\beta$ and predetermined $K$ topics. This is done $R$ times independently leading to a set of $N = R K$ topics in total. In our real text data example we make use of $K = 50$ topics and the total number of words used in that text corpus is $V = 25\,486$.
As in Section \ref{introduction}, $\bm z^{(r)} = (\bm z_1^{(r)}, ..., \bm z_K^{(r)}) \in \mathbb{N}_0^{V \times K}$ denotes the matrix of word counts per topic in the $r$-th replication. In addition, let 
\begin{align*}
n_k = \sum_{v=1}^V n_{k}^{(v)}
\end{align*}
be the total number of assignments to topic $\bm z_k$ and $\bm n = (n_1, ..., n_K )^T$ the corresponding vector of counts for topics $\bm z_1, ..., \bm z_K$. To take replications of modeling into account we use the notation
\begin{align*}
\bm z^{(\text{Rep})} = \left( \bm z^{(1)}, ..., \bm z^{(R)} \right) \in \mathbb{N}_0^{V \times N}.
\end{align*}
Then, for a given lower bound $\bm c = (c_1, ..., c_N)$ and for two topics $(i,j)$ represented by their word count vectors
\begin{align*}
\bm z_i, \bm z_j \in \left\{\bm z_1^{(1)}, ..., \bm z_K^{(1)}, \bm z_1^{(2)}, ..., \bm z_K^{(2)}, ..., \bm z_1^{(R)}, ..., \bm z_K^{(R)}\right\}
\end{align*}
our modified Jaccard coefficient is calculated by
\begin{align}
\label{modjaccard}
J_m(\bm z_{i}, \bm z_{j} \mid \bm c) := \frac{\sum\limits_{v = 1}^{V} \mathds{1}_{\left\{n_{i}^{(v)} > c_i ~\wedge~ n_{j}^{(v)} > c_j\right\}}\left(n_{i}^{(v)}, n_{j}^{(v)}\right)}{\sum\limits_{v = 1}^{V} \mathds{1}_{\left\{n_{i}^{(v)} > c_i ~\vee~ n_{j}^{(v)} > c_j\right\}}\left(n_{i}^{(v)}, n_{j}^{(v)}\right)}.
\end{align}
Reasonable choices for the threshold vector $\bm c = (c_1,\ldots, c_N)^T \in \mathbb{N}^N$ are an equal absolute lower bound ($c_i = c, i=1,\ldots,N$) for all words or a relative lower bound with $c_i = n_i/d, d \in \mathbb{N}$.

The interpretation of this modified Jaccard coefficient $J_m$ is the following. It is defined as the ratio of the numbers of the intersection and the union of the words of two topics, but a word is only considered, if the number of its occurrences in a text exceeds the threshold~$c$. In other words, we first restrict ourselves to the most relevant words per topic with respect to the number of assignments, in order to get rid of heavy tailed word lists. Then the resulting subsets of words are used to measure similarity of topics using the standard Jaccard coefficient.

We demonstrate how the measure is calculated with a small toy example. In \mbox{Table \ref{tab:counts}}, for eleven selected words the counts of assignments over all articles for the two topics $\bm z_1$ and $\bm z_2$ are given. We use the relative lower bound with $d=500$. In the analysis presented in Section \ref{results} we also use $d=500$, which leads to around 100 important words per topic. The last two columns indicate whether the corresponding word belongs to the modified intersection or union, respectively. For example, the word \textit{election} does not belong to the intersection because its count is below the topic specific (relative) threshold of at least nine assignments to topic $\bm z_1$.
\begin{table}[t]
\centering
\caption{Toy example: Assignment counts of two topics and calculation of modified Jaccard coefficient}
\label{tab:counts} 
\begin{minipage}{0.45\textwidth}
\begin{tabular}{rrrrr}
  \hline\noalign{\smallskip}
 & $\bm z_1$ & $\bm z_2$ & $\wedge$ & $\vee$ \\ 
  \noalign{\smallskip}\hline\noalign{\smallskip}
trump & 1\,668 & 2\,860 & 1 & 1 \\ 
  trumps & 446 & 854 & 1 & 1\\ 
  president & 91 & 876 & 1 & 1\\ 
  donald & 259 & 693 & 1 & 1\\ 
  news & 695 & 0 & 0 & 1\\ 
  said & 500 & 0 & 0 & 1\\ 
  election & 8 & 474 & 0 & 1\\ 
  will & 0 & 462 & 0 & 1\\ 
  women & 397 & 53 & 1 & 1\\ 
  debate & 394 & 11 & 0 & 1\\ 
  sarcastic & 1 & 4 & 0 & 0 \\ 
  \noalign{\smallskip}\hline\noalign{\smallskip}
  $\Sigma$ & 4\,459 & 6\,287 & 5 & 10 \\
  \noalign{\smallskip}\hline\noalign{\smallskip}
\end{tabular}
\end{minipage}
\begin{minipage}{0.42\textwidth}
\begin{align*}
\text{vocabulary size}~~ V &= 11,\\
\text{relative limit}~~ d &= 500\\
\Rightarrow \bm c &= \bm n/500\\
&= (4\,459, 6\,287)^T / 500 \\
&= (8.92, 12.57)^T.\\
\\
J_m(\bm z_1, \bm z_2 \mid \bm c) & = \frac{5}{10}.
\end{align*}
\end{minipage}
\end{table}
The ratio of the number of entries in the third and the fourth column results in the similarity $J_m(\bm z_1, \bm z_2 \mid \bm c) =\frac{5}{10}=0.5$ of the two given topics.

We choose the Jaccard coefficient with a slight modification as similarity measure for comparing topics based on their word count vectors. In the literature, several alternatives are discussed. \cite{aletras2014} argue that Jensen-Shannon Divergence is one of the best similarity measures based on word distributions considering correlation with human judgements. Moreover, they figured out that a standard Jaccard coefficient is able to realize higher correlations to human judgements than other common similarity measures on specific datasets. \cite{kim2011} showed that Jaccard coefficients perform on par with Jensen-Shannon Divergence \citep{lin1991} which is a symmetric version of the Kullback-Leibler Divergence \citep{kullback1951}, and outperform a number of other popular similarity measures like cosine similarity and Kullback-Leibler Divergence. To quantify the quality of the similarity measures they compare the negative log-likelihood of the model as an indicator how well the model explains the data. They swap the best matching topics from models of two time slices and interpret an increase of the negative log-likelihood as deficiency of the specific similarity measure. For our analysis we use the modified Jaccard coefficient as defined above. We prefer the Jaccard coefficient over Jensen-Shannon Divergence because of its flexibility given by the lower bound $\bm c$ and improved interpretability. 

\subsection{S-CLOP: a similarity measure for multiple sets of objects}
\label{mms}

We introduce the new similarity measure S-CLOP for comparing sets of objects. In the context of LDA models, the objects are topics that are represented by word count vectors, and pairwise distances between objects are calculated with the modified Jaccard coefficient $J_m$. However, the measure can be applied for general sets of objects with a corresponding distance respectively similarity measure.

The general idea of the measure S-CLOP is the following. First, join all sets to one overall set, then cluster the objects with subsequent local pruning, and then check how many members of the original different sets are contained in the resulting clusters. Then the disparity from the perfect situation of one representative from each set is calculated. 

For a set of topic models the objects are topics, and two models are very similar, if always one topic of the first model is clustered together with one topic from the other model. For the specific case of multiple LDA runs on the same data set (replications), a high similarity value means that many topics can be identified that have a representative in each LDA run.
In the following, we explain S-CLOP for the application with topic models, and typically use 50 topics per LDA run.

For the initial clustering step, we use hierarchical clustering with complete linkage \citep[pp. 520--525]{hastie2009}. We prefer complete linkage over single or average linkage because it uses the maximum distance between objects to identify clusters. This is consistent with our aim of identifying highly homogeneous groups. Of course, topic similarities must first be transformed to distances to apply hierarchical clustering.

\paragraph{Measuring disparity of a set of objects}

Consider a cluster (respectively a group) $g$ of topics, after clustering $R$ LDA runs in one joint cluster analysis, using all $R\cdot K$ topics from all runs. The goal is to quantify the deviation from the desired situation that each run is represented exactly once in $g$.
The vector $\bm t^{(g)} = (t_1^{(g)}, ..., t_R^{(g)})^T \in \mathbb{N}_0^R$ contains the number of topics that belong to the different LDA runs.

Then we define the disparity measure
\begin{align*}
U(g) := \frac{1}{R} \sum\limits_{r=1}^R \vert t_r^{(g)} - 1 \vert \cdot \sum\limits_{r=1}^R t_r^{(g)}.
\end{align*}
The first factor $\vert t_r^{(g)} - 1 \vert$ measures the deviation from the best case of exactly one topic per run in $g$. The second factor determines the number of members in the cluster and is required to penalize large clusters. Without this adjustment, the algorithm presented below for minimizing the sum of disparities would prefer one large cluster over a number of small clusters. In particular, without the second term, joining two perfect clusters as well as splitting one perfect cluster in two clusters would result in the same value for the mean disparity ($R/R=1$), and we prefer the second situation, where two different topics from one run are not clustered together. 
The disparity of one overall cluster $g$ containing all topics, e.g.\ defined by the root of a dendrogram, is given by $U(g) = (K-1)\cdot N$. 

\paragraph{Finding the best set of clusters by minimizing average disparity}

The goal is to minimize the sum of disparities $U(g)$ over all groups $g\in G$ of a cluster result. Hierarchical clustering of all $N$ objects (topics) provides a cluster result with $k$ clusters for all values of $k$ in $\{1,\ldots,N\}$.
A common approach then is to globally cut the dendrogram according to the target value~$k$. Here, we propose to prune the resulting dendrogram locally to obtain the final clusters. The pruning algorithm requires as input a hierarchical clustering result and minimizes the sum of disparities, with respect to the dendrogram structure, i.e.
\begin{align*}
U_{\Sigma}(G) := \sum_{g \in G} U(g) \to \min,  
\end{align*}
where $G$ is a set of clusters (of topics), and the set of all topics is a disjoint union of the members of the single clusters $g \in G$. 

Denote by $G^*$ the optimal set of clusters resulting from splits identified from the dendrogram, and by $U^* := U_{\Sigma}(G^*)$ the corresponding minimal sum of disparities.
The root of the dendrogram contains as a disjoint union the members of the two nodes obtained by the first split. Likewise, iteratively, each node contains as a disjoint union the members of the two nodes on a clustering level one step below this specific node, as denoted in Algorithm~\ref{pseudoU} by \textsf{node.left} and \textsf{node.right}.
The optimal sum $U^*$ can be calculated recursively with Algorithm~\ref{pseudoU}.

\begin{algorithm}[t]
\caption{Determining the minimal sum of disparities $U^*(g)$ of a cluster $g$}
\label{pseudoU}
 \KwData{A \textsf{node} of a dendrogram}
 \KwResult{The minimal possible sum of disparities for this node}
  \eIf{\em \textsf{is.leaf(node)}}{
   \KwRet{$(R-1)/R$}
   }{
   \KwRet{\em $\min\{U(\textsf{node}), \text{Recall}(\textsf{node.left}) + \text{Recall}(\textsf{node.right})\}$}
  }
\end{algorithm}

For a node in the dendrogram, we denote by $U(\textsf{node})$ the disparity of the corresponding cluster and by $U^*(\textsf{node})$ the minimal sum of disparities of the dendrogram induced by (or below) this node.
Algorithm \ref{pseudomms} can now be used to find the best set of clusters.
A cluster is added to the list of final clusters, if its disparity is lower than every sum of disparities obtained when further splitting this node.

\begin{algorithm}[t]
\caption{Finding the optimal set of clusters $G^*$}
\label{pseudomms}
 \KwData{A dendrogram with a \textsf{root}}
 \KwResult{A list correponding to the optimal set of clusters $G^*$, obtained by local pruning of the dendrogram}
 \Begin{
 $\textsf{node} = \textsf{root}$\;
  \eIf{\em $U(\textsf{node}) == U^*(\textsf{node})$}{
   Add all objects belonging to the cluster correponding to \textsf{node} as one cluster to the list\;
   }{
   Recall(\textsf{node.left})\;
   Recall(\textsf{node.right})\;
  }
  }
  \KwRet{\em list}
\end{algorithm}

\paragraph{Measuring stability with aggregated disparities}

Finally, we can calculate the similarity of a set of LDA runs using the optimized set of clusters. We normalize the sum of disparities of the optimal clustering, such that its values lie in the interval $[0,1]$, where $0$ corresponds to the worst case and $1$ to the best case. The worst case is a pruning state with $R$ clusters, each consisting of all topics from one LDA run. Then the pruning of Algorithm \ref{pseudomms} would lead to a set $\tilde{G}$ of $N$ single topic clusters, resulting in the highest possible value for the sum of disparities
\begin{align*}
U_{\Sigma,\textsf{max}} := \sum_{g \in \tilde{G}} U(g) = N \cdot \frac{R-1}{R}.
\end{align*}

The similarity measure S-CLOP ({\bf S}imilarity of multiple sets by {\bf C}lustering with {\bf LO}cal {\bf P}runing) for calculating the similarity of replicated LDA runs then is defined, for the identified optimal set of clusters $G^*$, by
\begin{align}
\label{mmsMeasure}
\text{S-CLOP}(G^*) := 1 - \frac{1}{U_{\Sigma,\textsf{max}}} \cdot \sum_{g \in G^*} U(g) ~\in [0,1].
\end{align}
Note that in the special case of comparing just two LDA runs with the same number of topics~$K$, like in the example in Section~3, the normalization factor is $U_{\Sigma,\textsf{max}} = 2\cdot K \cdot\frac{1}{2} = K$.

The introduced methods have been implemented as \textsf{R} package and are available at the GitHub repository \url{https://github.com/JonasRieger/ldaPrototype}. A release on \textsf{CRAN} is intended.

\section{Results}
\label{results}

In this chapter we present an analysis on a text corpus from the newspaper  \textit{USA Today}.
Latent Dirichlet Allocation (LDA) applied multiple times to this text corpus results in different word and topic distributions, depending on the initial random assignments. Thus, controlling the initialization would lead to more stable models, but the restriction on a subset of possible models can lead to less interpretable results. Instead, to improve the reliability, we aim to select out of a set of LDA runs a prototype that is most representative.

For modeling we use a Collapsed Gibbs Sampler implemented in \textsf{R 3.5.2} \citep{R} in the package \textsf{lda} \citep{lda}. We run the Sampler for 270 iterations. Further, we make use of a number of other \textsf{R} packages. We use \textsf{batchtools} \citep{batchtools} to run the modeling procedure on a batch system, and \textsf{data.table} \citep{data.table} for an improved implementation of data frames in \textsf{R}. The packages \textsf{dendextend} \citep{dendextend} and \textsf{RColorBrewer} \citep{RColorBrewer} are useful for handling and displaying dendrograms, while preprocessing of the text data is done with routines from \textsf{tm} \citep{tm, tm.paper} and \textsf{tosca} \citep{tosca}.

\subsection{Data}
\label{data}
As an example corpus we introduce a set of articles published in the daily newspaper \textit{USA Today} in an interval of six months, from June until November 2016. The corpus consists of $M = 7\,657$ articles. The dataset is provided by \cite{nexis} and is preprocessed with common procedures in natural language processing (NLP). Duplicates from articles that occur more than once are removed, so that every unique article remains once. This step excluded 204 articles from the analysis. As common in practice, characters are formatted to lowercase, numbers and punctuation are removed. In addition, a trusted stopword list \citep{snowball} is applied to remove words that do not help in classifying texts in topics. Moreover, the texts are tokenized and words with a total count less than six are neglected. This reduces the vocabulary size from 79\,734 to $V = 25\,486$. For our example corpus we model $K = 50$ topics per LDA and we assume $\alpha$ and $\beta$ to be $1/50$.

\subsection{Cluster analysis and similarity calculation}
\label{example}
We run LDA four times, such that the number of runs to be compared is $R = 4$ and the total number of topics to be clustered is $N = R\cdot K = 4\cdot 50 = 200$. To demonstrate how dissimilar replicated LDA runs can be, we cluster the $N = 200$ topics from the $R = 4$ independent runs with $K = 50$ topics each using the modified Jaccard coefficient $J_m$ from (\ref{modjaccard}), complete linkage and the new algorithm for pruning. The four runs were selected from $10\,000$ total runs. The selection criterion is described in Section \ref{increase}.

We apply hierarchical clustering with complete linkage to the $200$ topics. The topics are labeled with meaningful titles (words or phrases). These labels were obtained by hand, based on the ranked list of the 20 most important words per topic. For this, the importance of a word $v = 1, ..., V$ in topic $k = 1, ..., K$ \citep{lda} is calculated by
\begin{align*}
I(v, k) = \frac{n_k^{(v)}}{n_k} \cdot \left[ \log \left(\frac{n_k^{(v)}}{n_k}+\varepsilon\right) - \frac{1}{K} \cdot \sum_{l=1}^K \log \left(\frac{n_l^{(v)}}{n_l}+\varepsilon\right) \right],
\end{align*}
where $\varepsilon$ is a small constant value which ensures numerical computability, here we use $\varepsilon=10^{-5}$. The importance measure is intuitive, because it scores words high which occur often in the present topic, but less often in average in all other topics.

\begin{figure}
	\includegraphics[height = \textwidth, angle = -90]{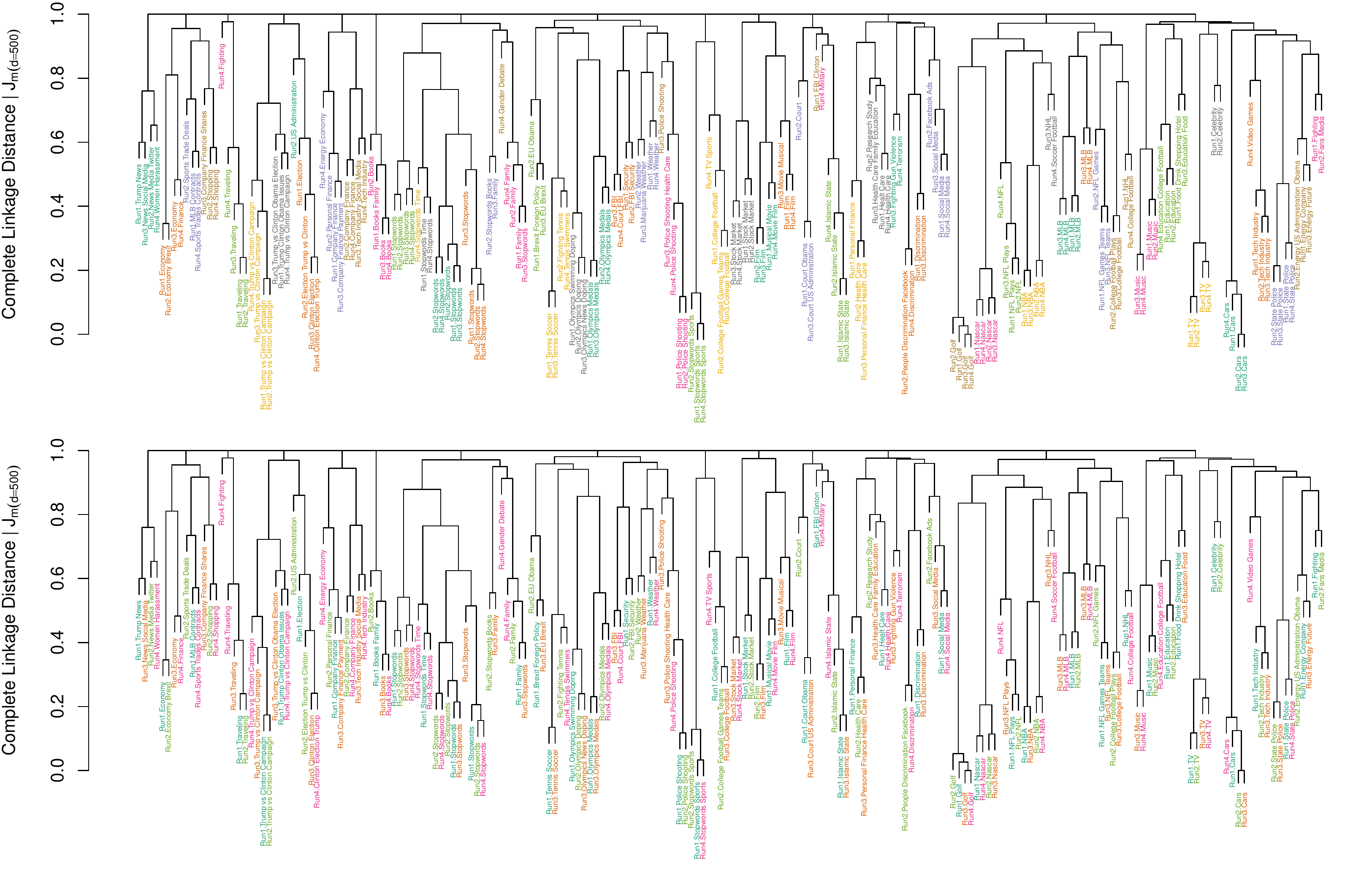}
	\caption{Dendrograms of $N = 200$ topics from $R = 4$ selected LDA runs with $K = 50$ topics each; left: colored by runs; right: colored by cluster membership}
	\label{fig:dend}
\end{figure}

Figure~\ref{fig:dend} shows two dendrograms visualizing the result of clustering the 200 topics and of Algorithm \ref{pseudomms} for clustering with local pruning. The horizontal axis describes the complete linkage distance based on our modified Jaccard coefficient $J_m$ with $d=500$, which was used to cluster topics. In the left dendrogram, the topic labels are colored with respect to the LDA run (\textit{Run1}: petrol, \textit{Run2}: green, \textit{Run3}: orange, \textit{Run4}: red).
In the right dendrogram, topic labels are colored according to the clusters obtained with our proposed pruning algorithm. In addition, every topic label is prefixed by its run number.

Looking at the right dendrogram, we see that often the titles of topics in the identified topic clusters are very similar. This demonstrates that the four LDA runs often produce similar topics that are represented by similar word distributions. Examples for such stable topics are \textit{Trump vs Clinton Campaign} colored yellow and \textit{Olympics Medals} colored green.

However, there are also considerable differences visible. In the left dendrogram in Figure \ref{fig:dend}, strikingly, there are several topics from \textit{Run4}, highlighted in red, where no other topic is within a small distance. It is remarkable that \textit{Run4} creates such a number of individual topics, e.g. \textit{Video Games}, \textit{Gender Debate}, \textit{TV Sports} (which includes words for describing television schedules of sport events) and \textit{Terrorism}. Also, \textit{Run4} leads to six explicit stopword topics, the maximum number compared to the other runs with four to six stopword topics.

In the right dendrogram, the color depends on cluster membership. We measure combined stability of these four LDAs by applying the proposed pruning algorithm (Algorithm \ref{pseudomms}) to the dendrogram. This leads to 61 clusters and a stability of $0.83$. The normalization factor is given by $U_{\Sigma,\textsf{max}} = K \cdot (R-1) = 50 \cdot 3 = 150$, and the minimization of the sum of disparities yields $U^* := U_{\Sigma}(G^*) = 25$, resulting in the similarity $\text{S-CLOP} = 1-25/150 = 0.83$. There are seven single topics, one from each of the first three runs and four from \textit{Run4}. The eleven clusters which consist of exactly three topics contain ten times a topic from \textit{Run1}. Topics from \textit{Run2} and \textit{Run3} are represented nine times each, whereas only five of the mentioned clusters contain a topic from \textit{Run4}. This shows that LDA run \textit{Run4} strongly differs from the others. In many cases only a topic from this run is missing to obtain perfect topic clusters.

\subsection{Increase of reliability}
\label{increase}

Finally, we demonstrate how to determine a prototype LDA run as the most representative run out of a set of runs, based on our novel pruning algorithm. We show that this technique leads to systematically higher LDA similarities, which suggests a higher reliability of LDA findings from such a prototype run.

The stability measure S-CLOP in (\ref{mmsMeasure}) quantifies pairwise similarity of two LDA runs by
\begin{align*}
1 - \frac{1}{50} \cdot \sum_{g \in G^*} U(g),
\end{align*}
where $K = 50$ is the number of topics per model and $G^*$ an optimized set of topic clusters identified by our proposed pruning algorithm. We investigate the stability measure S-CLOP on the corpus from \textit{USA Today}.

\begin{figure}
\includegraphics[width = \textwidth]{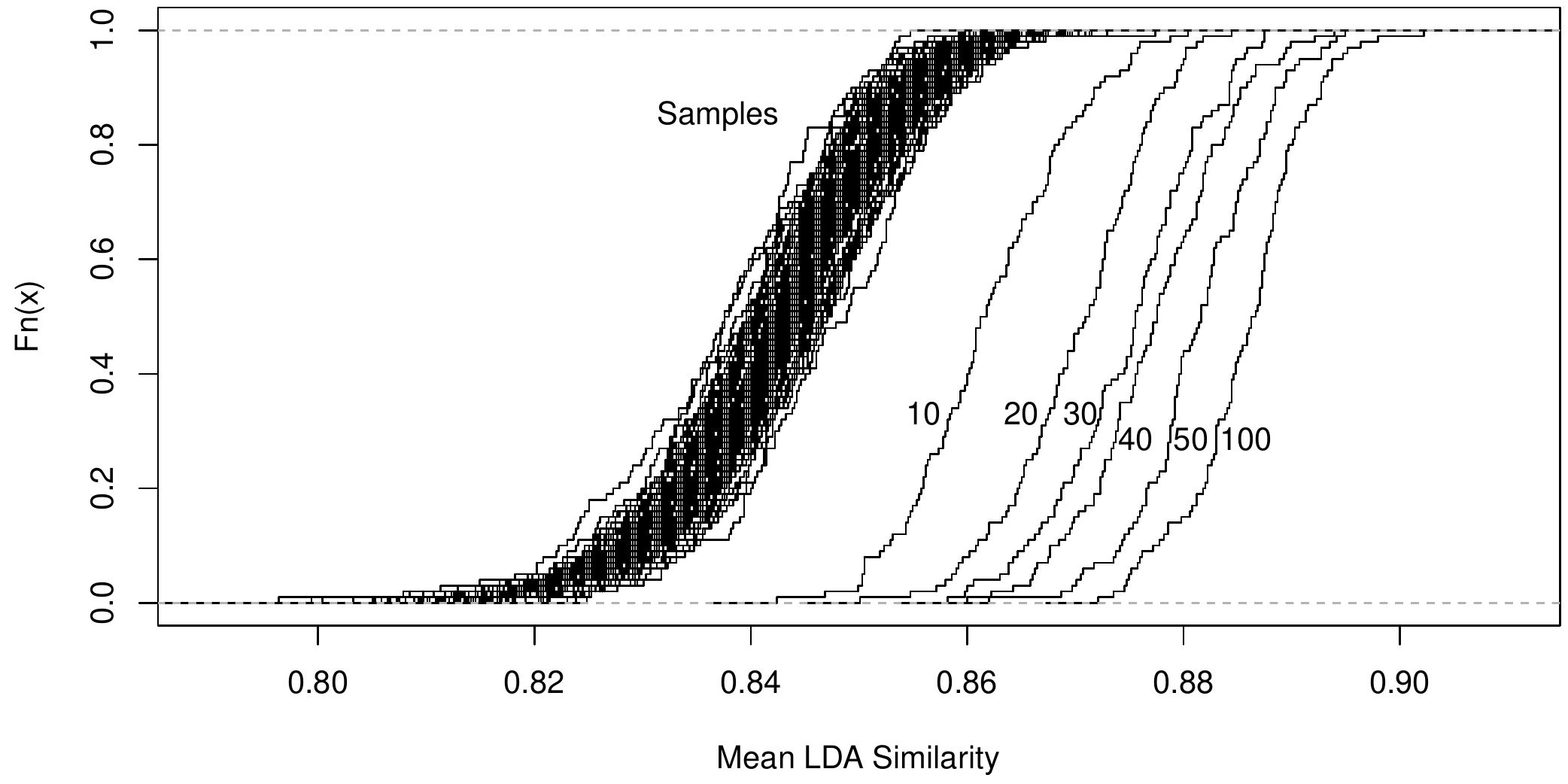}
\caption{Empirical cumulative distribution functions of mean similarities calculated (i) on 100 samples of randomly selected LDA runs (left, labeled with Samples) and (ii) on the 100 most representative prototype LDA runs based on (sub)samples of $10, 20, 30, 40, 50$ or all 100 LDA runs}
\label{fig:stability}
\end{figure}

We propose to select the LDA run with highest mean pairwise similarity (measured with S-CLOP) to all other runs. The following study shows that this is a suitable way to identify a stable prototype LDA, thus leading to improved reliability of LDA findings based on this particular run. We fit 100 LDA models and select the model with highest mean similarity as prototype. This procedure is repeated 100 times, which results in 100 prototype models. Then, for the 100 prototypes, also mean pairwise similarities to the other prototypes are calculated. The results are visualized in Figure~\ref{fig:stability}.
The very right curve describes the empirical cumulative distribution function of the mean similarities obtained for the 100 prototypes. At the very left there are the 100 curves of the $100 \times 100$ original runs. In addition, we also determine 100 prototypes from subsamples. For this, only 10, 20, 30, 40 or 50 runs from each original set of 100 runs are randomly selected and are used for the following calculation steps. The resulting curves are also plotted and labeled in Figure~\ref{fig:stability}.

The minimum of the mean similarities from the original 100 sets of 100 models is $0.796$, while the maximum is $0.877$. We recommend to fit at least 50 replications because this leads to an increase of similarity to $0.862$ at the minimum and $0.895$ at the maximum. Higher values for the number of repetitions are desirable. In general, the choice depends on the complexity of the corpus. Encapsulated topics or certain complicated dependency structures make the modeling procedure more prone to a larger span of possible fits and therefore to smaller mean similarity values. However, if computational power is limited, already taking the prototype model from 10 candidates considerably improves the stability. Here, the minimum and maximum of mean similarity are $0.842$ and $0.880$, considerably higher values than without replications.

The dendrogram in Figure \ref{fig:dend} (in Section \ref{example}) illustrates that random selection can lead to a poor model regarding interpretability. In fact, the runs \textit{Run1} and \textit{Run2} were chosen as the top two models in mean similarity of the 100 prototypes from samples of size 100, which means their points lie at the top of the very right curve in Figure~\ref{fig:stability}. Their similarity values are $0.902$ and $0.898$ in the set of prototypes or $0.877$ and $0.871$ in the original sets, respectively. The model \textit{Run3} was chosen as the worst of the 100 prototype models with a similarity value of $0.872$ and $0.863$ in its original set. \textit{Run4} was chosen randomly as one of the worst models realizing a mean similarity to all other models in its original set of $0.807$. Thus our results of the reliability analysis in Figure \ref{fig:stability} show that random selection can lead to low reliability. Instead, we recommend to use the replication and prototype approach to increase mean similarity, which comes along with an improvement in reliability.

\section{Discussion}
\label{discussion}

Topic modeling is popular for understanding text data, though the analysis of the reliability of topic models is rarely part of applications. This is caused by plenty of possibilities for measuring reliability, but missing strategies for increasing reliability without touching the original fitting procedure.

We introduce a novel algorithm for assessing the stability of LDA by calculating pairwise similarities of replicated runs and quantifying similarity of sets of runs with our new measure S-CLOP. The ideal situation is that clustering topics from different runs leads to groups of topics that consist of exactly one topic per run. We show that two random selected models are likely to produce considerably different topic structures. To overcome this issue, we propose a method for increasing similarity by selecting prototypes from repeated LDA runs.
Then, also reliability of the prototypes is higher than reliability of the original runs. The improvement is reached by repeating the modeling procedure and selecting the prototype model that is on average most similar to all other runs, using our introduced stability measure. We recommend to fit at least 50 models depending on the corpus complexity. 
The idea to use replicated LDA runs and select a prototype can be transferred to other similarity measures than the modified Jaccard coefficient $J_m$ that was used here.
We make use of a modified Jaccard coefficient because of its flexibility and interpretability.
In addition, our method is also usable to compare replications by matching topics of other topic models than LDA.

It is of interest to generalize our new stability measure S-CLOP to a similarity measure for models based on different text corpora. There are several difficulties to consider. For example, it is an open question how to handle differences in the number of topics $K$ of compared models. In addition, it needs to be analyzed whether a comparison of a number of runs per corpus or a comparison of prototypes is more practical. Such similarity measures as the one proposed here can open various fields of other applications. For example, similarities in reporting of newspaper offices or differences in coverage on various media channels like twitter, online and print can be quantified.


\bibliographystyle{spbasic}
\bibliography{literatur}   

%
%

\end{document}